\documentclass[times,twocolumn,final]{elsarticle}

\usepackage{lmodern}
\usepackage{medima}
\usepackage{multirow}

\usepackage{amssymb}
\usepackage{latexsym}
\usepackage{float}
\usepackage[caption = false]{subfig}
\usepackage{url}
\usepackage{xcolor}

\definecolor{newcolor}{rgb}{.8,.349,.1}

\journal{Medical Image Analysis}

\begin{document}

\verso{M. Kang \textit{et~al.}}

\begin{frontmatter}

\title{Dual-stream Pyramid Registration Network\tnoteref{tnote1}}%
\tnotetext[tnote1]{This work was initially presented at MICCIA 2019.}

\author[1,2]{Miao Kang}

\author[1]{Xiaojun Hu}

%\fntext[fn1]{This is author footnote for second author.}
\author[1]{Weilin Huang\corref{cor1}}
\cortext[cor1]{Corresponding author: whuang@malongtech.com}

%% Third author's email
%\ead{author3@author.com}
\author[1]{Matthew R. Scott}
\author[3]{Mauricio Reyes}

\address[1]{Malong LLC, Wilmington, USA}
\address[2]{Institute of Artificial Intelligence and Robotics, Xi’an Jiaotong University}
\address[3]{ARTORG Center for Biomedical Engineering Research, Univ. of Bern, Switzerland}

%\received{1 May 2013}
%\finalform{10 May 2013}
%\accepted{13 May 2013}
%\availableonline{15 May 2013}
%\communicated{S. Sarkar}

\begin{abstract}
%%%
We propose a Dual-stream Pyramid Registration Network (referred as Dual-PRNet) for unsupervised 3D brain image registration.
Unlike recent CNN-based registration approaches, such as VoxelMorph, which computes a registration field from a pair of 3D volumes using a single-stream network, we design a two-stream architecture able to estimate multi-level registration fields sequentially from a pair of feature pyramids.
%
%we propose a two-stream architecture that computes two convolutional feature pyramids separately. We design a pyramid registration module to predict multi-scale registration fields directly from the computed feature pyramids.
%
Our main contributions are: (i) we design a two-stream 3D encoder-decoder network that computes two
convolutional feature pyramids separately from two input volumes;
%
%generating strong deep representations for deformation estimation;
(ii) we propose sequential pyramid registration where a sequence of  pyramid registration (PR) modules is designed to predict multi-level registration fields directly from the decoding feature pyramids.  The registration fields are refined gradually in a coarse-to-fine manner via sequential warping, which equips the model with a strong capability for handling large deformations;
(iii) the PR modules can be further enhanced by computing local 3D correlations between the feature pyramids, resulting in the improved Dual-PRNet$^{++}$ able to aggregate rich detailed anatomical structure of the brain;
(iv) our Dual-PRNet$^{++}$ can be integrated into a 3D segmentation framework for joint registration and segmentation, by precisely warping voxel-level annotations. Our methods are evaluated on two standard benchmarks for brain MRI registration, where Dual-PRNet$^{++}$ outperforms the state-of-the-art approaches by a large margin, i.e., improving recent VoxelMorph from 0.511 to 0.748 (Dice score) on the Mindboggle101 dataset. In addition, we further demonstrate that our methods can greatly facilitate the segmentation task in a joint learning framework, by leveraging limited annotations. Our code is available at: \url{https://github.com/kangmiao15/Dual-Stream-PRNet-Plus}.

%, e.g., having improvements over recent VoxelMorph [2] with 0.683->0.778 on the LPBA40, and 0.511->0.631 on the Mindboggle101, in term of average Dice score.

%Based on the preliminary architecture, in this work, we futher develop the correlation representations for deformation field estimation and design a more effective architecture of pyramid registration module. we also utilize Dual-PRNet$^{++}$ to provide auxiliary labeled data for segmentation module, which improve the segmentation performance greatly. Our contributions are three-fold: (i)we design a 3D correlation layer to compute match cost of corresponding patch of a pair of  convolutional features, which provide the explicit correlation information for the estimation of deformation field.(ii) we add a feature fusion block on pyramid registration module to enforce the network to exploit the implicit correspondence in feature space. The concatnate of explicit and implicit  features boosts the performance significantly with few parameter increased. (iii)we extend our work on the integration framwork of segmentation and registration, which indicate that our Dual-PRNet$^{++}$ can provide higher quantified labeled data to enhance the performance of segmentation task. The enhanced Dual-PRNet is evaluated on the standard benchmarks with more detail structures for brain MRI registration, where it outperforms the previous approaches by a large margin, e.g., having improvements over Dual-PRNet with 0.631$\rightarrow$0.748 on the Mindboggle101, in term of average Dice score.
%%%%
\end{abstract}

\begin{keyword}
%% Keywords
\KWD Medical image registration\sep encoder-decoder network\sep deformable registration\sep 3D segmentation\sep brain MRI
\end{keyword}

\end{frontmatter}

%\linenumbers

\section{Introduction}
\label{sec1}
Deformable image registration has been widely used in image diagnostics, disease monitoring, and surgical navigation, with the goal of learning the anatomical correspondence between a moving image and a fixed image. A registration process mainly consists of three steps: establishing a deformation model, designing a function for similarity measurement, and a learning step for parameter optimization. Traditional deformable registration methods, such as Demons~\citep{vercauteren2009}, Large Diffeomorphic Distance Metric Mapping (LDDMM)~\citep{glaunes2008} and symmetric
normalization (SyN)~\citep{avants2008}, often cast the deformable registration as a complex optimization problem that involves intensive computation by densely measuring voxel-level similarities.
Recent deep learning technologies have advanced this task  considerably  by developing learning-based approaches, which allow them to leverage the strong feature learning capability of deep  networks  \citep{Miao2016, 2017Quicksilver, hering2019, hering2019Memory, nielsen2019, Kuckertz2020},  resulting in fast training and accurate inference, e.g., by taking orders of magnitude less time.

 However, learning-based approaches for medical image registration often require strong supervised information, such as ground-truth registration fields or anatomical landmarks.  
While obtaining a large-scale medical dataset with such strong annotations is extremely expensive, which inevitably limits the clinical application of supervised approaches.
Recently, unsupervised learning-based registration methods have been developed, by learning a registration function that maximizes the similarity between a moving volume and a fixed volume.
For example, Balakrishnan \emph{et al.} proposed VoxelMorph able to learn a parameterized registration function using a convolutional neural network (CNN) \citep{balakrishnan2018}.  Furthermore, they introduced an auxiliary loss able to integrate segmentation masks into the loss function, as described in~\citep{balakrishnan2019}. 
%VoxelMorph estimates a deformation field by using an encoder-decoder CNN, and then warps the moving image with a spatial transformation layer~\citep{jaderberg2015}.
%Furthermore, Kuang and Schmah developed an unsupervised method, named as FAIM, which extends VoxelMorph with a new penalty loss on negative Jacobian determinants \citep{kuang2019}.
%
However, Lewis \emph{et al.} demonstrated that the performance of existing CNN-based approaches can be limited in real-world clinical applications, where two medical images or volumes may have significant spatial displacements or large slice spaces~\citep{lewis2018}.

Recent approaches on optical flow estimation attempted to handle large displacements by gradually refining the estimated flows~\citep{ranjan2017, hui2018}. For example, Ranjan \emph{et al.} estimated multi-resolution optical flows with a Spatial Pyramid Network (SPN) to warp a moving volume at each pyramid level. Hui \emph{et al.} introduced feature warping to replace image warping in the process of pyramidal feature refinement, resulting in a lightweight yet effective network. 
 More recently, Eppenhof  \emph{et al.}  attempted to train neural networks progressively to handle the problem of large displacements, by expanding the networks gradually with additional layers that are trained on higher resolution data \citep{eppenhof2019progressively}.   
This inspired us to design a sequential warping mechanism able to warp two volumes gradually in a coarse-to-fine manner.
In addition to learning meaningful feature representation, medical image registration also requires strong pixel-wise correspondences between moving and fixed volumes, which naturally involves learning local correlations between intermediate features of the moving and fixed volumes. Therefore, current optical flow estimation methods, such as \citep{dosovitskiy2015, sun2018, hui2018},
%which perform the similar task with registration,
utilize a correlation layer to enable the network to identify the actual correspondences from convolutional features \citep{dosovitskiy2015}.
This also inspired us to develop a new 3D correlation layer capable of learning such correlations to further enhance feature representation.

In addition, our registration network is able to  capture the semantic correspondence between moving and fixed volumes. This allows it to accurately warp the anatomical annotations of moving volumes to the fixed volumes, providing rough supervised information for training a segmentation network on the target volumes where the annotations are not available \citep{estienne2019, zhaoamy2019, hu2019}. Recent work in \citep{xu2019} shows that such warped labels can be used as auxiliary data to improve the performance of segmentation when the training data with annotations is very limited. 

Furthermore, Estienne \emph{et al.} proposed an U-ReSNet \citep{estienne2019} which is a lightweight framework for joint registration and segmentation, with excellent results achieved. This further confirmed the benefits of joint learning.
%The promising results of U-ReSNet \citep{estienne2019}, which is a lightweight joint framework for registration and segmentation, also prove the benefits of task combination. 
In addition, Wang \emph{et al.} presented a label transfer network (LT-Net) able to propagate a segmentation map from the atlas to unlabelled images, by learning the reversible voxel-wise correspondences \citep{wang2020lt}.   In this work, we demonstrate that our Dual-PRNet$^{++}$ can be integrated into a 3D segmentation framework for joint segmentation and registration, which facilitates the segmentation task using limited manual annotations.

\begin{figure*}[htbp]
	\centering
	\includegraphics[width=18cm,height=6cm]{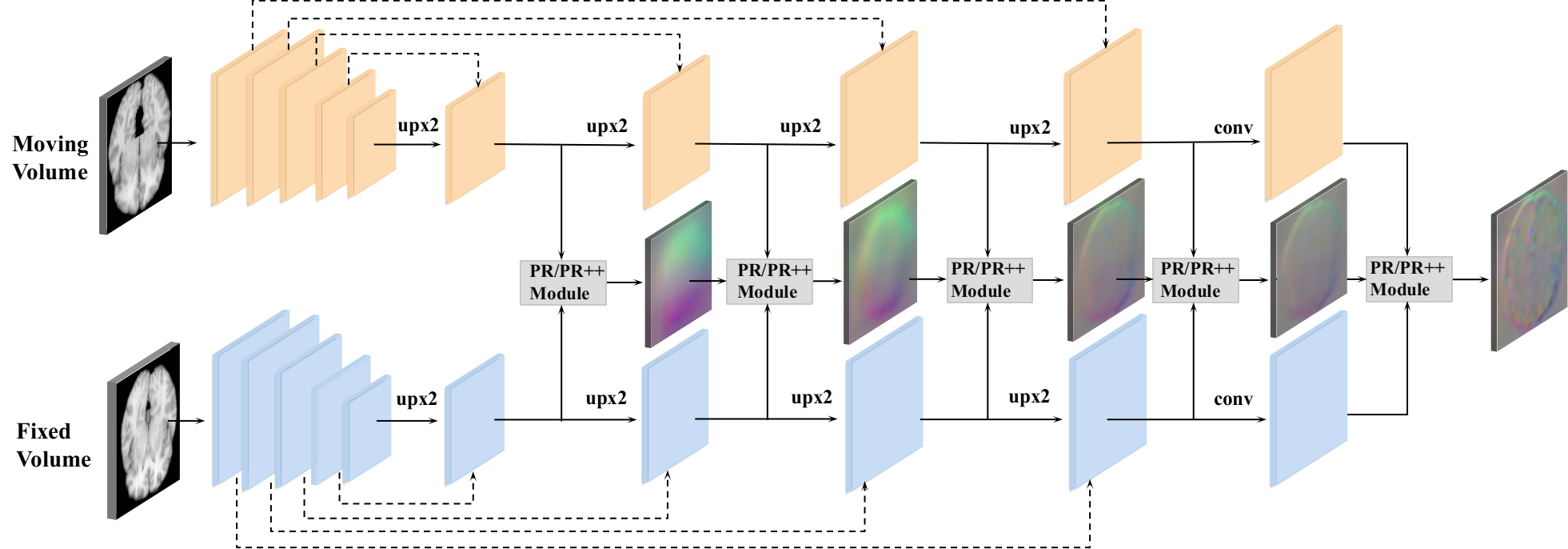}
	%\vspace{-4mm}
	\caption{Framework of the proposed Dual-PRNet$^{++}$, which is a dual-stream encoder-decoder network, integrated with new sequential pyramid registration including a sequence of pyramid registration (PR) modules or PR$^{++}$ modules. The dual-stream model computes two convolutional feature pyramids separately from two input volumes, while the PR / PR$^{++}$ modules estimate a sequence of deformation fields which can warp the pyramid features gradually in a coarse-to-fine manner. Finally, the final deformation field is generated by sequentially
warping the estimated fields, as shown in Fig. \ref{fig:final_DF}.
	} \label{fig:dual_prnet}
\end{figure*}

\textbf{Contributions.}
This paper extends our preliminary version of Dual-PRNet presented at the Medical Image Computing and Computer Assisted Intervention (MICCAI) 2019 conference \citep{hukang2019}, with two main extensions: we introduce Dual-PRNet$^{++}$ by improving sequential pyramid registration (PR) with the enhanced PR$^{++}$ modules which boost the performance; and we apply our Dual-PRNet$^{++}$ for joint 3D segmentation and registration. The overall contributions can be summarized as:
(i) a two-stream 3D encoder-decoder network is designed to compute two
convolutional feature pyramids separately from two input volumes,
generating stronger deep features for deformation estimation;
(ii) we propose sequential pyramid registration where a sequence of registration fields is estimated by a set of designed pyramid registration (PR) modules. The estimated registration fields   perform  sequential warping over the decoding layers, which refines  the feature pyramids gradually in a coarse-to-fine manner. This equips the model with a strong capability for handling large deformations;
%
%can predict a sequence of registration fields directly from the decoding feature pyramids.  The registration fields are refined gradually in a coarse-to-fine manner via sequential warping, which enable the model with strong capability for handling significant deformations;
%
(iii) the PR module can be further enhanced by computing local 3D correlations (between two feature pyramids) followed by multiple residual convolutions, which aggregates  richer local details of anatomical structure for better estimating the deformation fields, resulting in the improved Dual-PRNet$^{++}$.  In addition, the 3D correlations with more convolutional layers in the PR$^{++}$ module are able to enlarge receptive fields which further enhance the ability to handle large deformations; 
(iv) our registration networks can be integrated into a 3D segmentation network, resulting in a unified 3D framework for joint segmentation and registration.
Finally, our methods are evaluated on brain MRI registration, where the Dual-PRNet$^{++}$ outperforms the state-of-the-art approaches by a large margin. In addition, on 3D segmentation with limited annotations, we demonstrate that our methods can greatly facilitate the segmentation task via joint framework, by accurately warping voxel-level annotations.

%three new technical improvements, which are the key to boost the performance.
%$\left.1\right)$ we develop a 3D correlation layer to compute the match cost on the corresponding patches of two input volumes, which explicitly compute strong correlation information for estimating the deformation fields;
%$\left.2\right)$ we propose a feature fusion block within the pyramid registration module, and at the same time, simplify the dual-stream encoder, forcing the network to learn the implicit corresponding from the feature space. This results in stronger deep representations for estimating multi-level deformations. With the integrated of correlation features and aggregate features, our new Dual-PRNet$^{++}$ can achieve a significant higher registration accuracy.
% $\left.3\right)$ we utilize our registration network to provide auxiliary voxel-level labels for segmentation task which can improve the performance significantly in an unsupervised manner.

\section{Related Work}
In this section, we briefly review recent approaches on learning-based medical image registration, particularly on using deep learning methods. More comprehensive studies on this topic can be referred to \citep{Boveiri2020, Fu2020, Haskins2020Deep}.

 Deep learning technologies have recently been applied to medical image registration. For example, Hu \emph{et al.} explored the strong capability of CNN to learn deformable image registration, with promising results achieved \citep{hu2017learning}.
%The strong capability of CNNs for learning deep representation have been integrated into learning deformable image registration, with promising results achieved \citep{hu2017learning}.
%Integrating CNN's strong capability of feature representation, the idea of learning registration has shown to be promising\citep{hu2017learning}.
In \citep{Miao2016}, CNN regressors were employed to directly estimate transformation parameters, while De Vos \emph{et al.}  attempted to develop a patch-based end-to-end unsupervised deformable image registration network (DIRNet) \citep{deVos2017}, where a spatial transformer network (STN) \citep{jaderberg2015} was applied for estimating a deformation field.
%
%by back-propagating an image similarity metric as a loss. 
%With the introduction of spatial transformer network (STN) \citep{jaderberg2015}
However, the deformation field estimated by STN is unconstrained, which may cause  severe  distortions. To overcome this limitation, VoxelMorph \citep{balakrishnan2018,balakrishnan2019} was proposed. It estimates a deformation field by using an encoder-decoder CNN with a regularization penalty on the deformation field. Furthermore, Kuang and Schmah developed an unsupervised method, named as FAIM, which extends VoxelMorph by introducing an explicit penalty loss computing negative Jacobian determinants \citep{kuang2019}.

 However, these methods may fail to estimate large displacements in complex deformation fields, and recent efforts have been devoted to handle this issue by developing stacked multiple networks \citep{de2019deep, zhao2019, KIM2021102036}.
%or multi-resolution strategies such as \citep {sokooti2017nonrigid, hering2019, mok2020large, liu2019}.
%
 
%Our work is also related to recent approaches which estimate t issue.
%fields at multiple scales. \cite{zhao2019,hering2019,liu2019}.
For example, Zhao \emph{et al.} designed recursive cascaded networks where multiple VoxelMorph are cascaded recursively, which were employed to warp the images gradually \citep{zhao2019}. Kim \emph{et. al} proposed CycleMorph, which consists of two registration networks, taking inputs by switching their orders with a cycle consistency. 
It can be extended to multi-scale implementation performing on large volumes. This allows the model to better capture transformation relationships at different levels, but at the cost of a high complexity and computational burden as it requires multiple models. 
In \citep{de2019deep}, multiple ConvNets were stacked into a larger architecture to perform image warping in a coarse-to-fine manner. 
More recently, several attempts have been made by cascading an affine alignment subnetwork and a deformable subnetwork to improve the performance \citep{zhu2020unsupervised, de2019deep, HuangWeijian2021a, Shengyu2020unsupervised}. 
%Recent methods, such as \citep{zhu2020unsupervised, de2019deep, HuangWeijian2021a, Shengyu2020unsupervised}, cascaded an affine alignment subnetwork and a deformable subnetwork for improving the performance. 
However, sequential combination of multiple networks will result in an accumulation of interpolation artifacts, which may affect the quality of the estimated deformation field.

Therefore, recent approaches attempted to estimate deformation fields at multiple resolutions \citep{sokooti2017nonrigid, hering2019, mok2020large, liu2019,jiang2020multi, lei20204d, risheng2021learning}. For example, a RegNet was introduced in \citep{sokooti2017nonrigid}, which can be trained by using a large set of artificially generated displacement vector fields (DVF), and then the feature maps computed at multiple scales are concatenated to equip the network with fusion information. 
In \citep{hering2019}, mlVIRNET was introduced by creating an image pyramid (not feature pyramid), where a single-stream network is applied multiple times for computing the deformation fields at different image resolutions.

Furthermore, Mok \emph{et. al} proposed a L-level Laplacian pyramid framework (named as LapIRN) to mimic the conventional multi-resolution strategy, which warps the images from the previous level \citep{mok2020large}. 
%
%LapIRN \citep{mok2020large} uses a L-level Laplacian pyramid framework to mimic the conventional multi-resolution strategy, and warps the images from the previous level. 
%
Eppenhof \emph{et al.}  attempted to  expand the networks progressively with  additional  layers  that  are  trained  on  higher  resolution  data \citep{eppenhof2019progressively}, and a final deformation field can be estimated by averaging multi-resolution deformation fields computed from the pyramidal structure of a U-Net \citep{cciccek20163d}.  %\citep{eppenhof2019progressively}.

These approaches commonly stack the moving volume and fixed volume together as the input of a single-stream CNN, which largely discards transformation relationships between the two volumes. Two-stream encoders have recently been developed, which are able to encode the two volumes separately for better aggregating multi-level correlations in the feature spaces. 
For instance, 
%Furthermore, we firstly integrate a volume change control term into the loss function of a deep learning-based registration method to penalize occurring foldings inside the deformation field
%
%For instance,   and kuckertz  \emph{et al.} \citep{Kuckertz2020} utilized a dual-stream architecture on the highest resolution in order to learn specific features from different modality images.  
Krebs \emph{et al.} proposed an efficient latent variable model, which maps similar deformations close to each other in an encoding space \citep{krebs2019learning},
 while Hering \emph{et al.} developed a 2.5D two-stream convolutional transformer architecture, which is a memory-efficient weakly supervised learning model for multi-modal image registration \citep{hering2019Memory}. 
In \citep{liu2019}, a dual-stream network was developed to predict multi-resolution deformation fields from different convolutional layers independently, which are then enlarged and averaged to generate a final deformation field. 
Similarly, Liu \emph{et al.} utilized a dual-stream encoder to obtain two feature pyramids, and then computed a single transformation field with a contrastive loss and a single-stream decoder \citep{Lihao2020}. 
 Besides, the two-steam design was further applied in \citep{Kuckertz2020}  where  two generators with U-Net
architecture and two discriminators using patchGAN \citep{Isola2017}  were developed.

In this paper, we design a dual-stream network to compute two meaningful feature pyramids separately, and directly estimate sequential deformation fields in the feature space, in a single pass. Refinements on both registration fields and convolutional features are performed in a layer-wise, sequential, and coarse-to-fine manner, providing an efficient approach to align the two volumes gradually and more accurately in the feature space. This results in an end-to-end trainable model for unsupervised 3D image registration.  
%which is  adopted by recent work \citep{zhou2020unsupervised}.

\section{Dual-Stream Pyramid Registration Network}
In this section, we describe the details of the proposed Dual-PRNet and Dual-PRNet$^{++}$, including three main components: (i) a dual-stream encoder-decoder network for computing feature pyramids, (ii) sequential pyramid registration, and (iii) the improved pyramid registration (PR) modules: PR$^{++}$ modules.

\subsection{Preliminaries}
The goal of 3D medical image registration is to estimate a deformation field $ \Phi $ which can warp a moving volume  $M\subset R^{H\times W\times D}$  to a fixed volume  $ F\subset R^{H\times W\times D}$ , so that the warped volume  $ W=M\circ\Phi\subset R^{H\times W\times D}$  can be accurately aligned to the fixed one $ F $. We use $M\circ\Phi$ to denote the application of a deformation field $ \Phi $ to the moving volume with a warping operation, with image registration being formulated as an optimization problem:
\begin{equation}
\hat{\Phi} = \arg \min_{\Phi} \mathcal{L}(F,M,\Phi)
\end{equation}
\begin{equation}
\mathcal{L}(F,M,\Phi)= \mathcal{L}_{sim}(F, M\circ\Phi)+\lambda\mathcal{L}_{smooth}(\Phi)
\end{equation}
where $\mathcal{L}_{sim}$ is a function measuring the image similarity between the warped image ($M \circ \Phi$) and the fixed image ($F$), and $\mathcal{L}_{smooth}$ is a regularization constraint on the deformation field ($\Phi$), which enforces spatial smoothness. Both $\mathcal{L}_{sim}$ and $\mathcal{L}_{smooth}$ can be defined in various forms,  and recent efforts have been devoted to developing a powerful approach to computing the deformation field $ \Phi $.  For example, VoxelMorph~\citep{balakrishnan2018, balakrishnan2019} uses a CNN to compute a deformation field, $\Phi=f_{\theta}(F,M)$, where $\theta$ are learnable parameters of the CNN. In VoxelMorph,
the deformation warping operation is implemented by using a spatial transformer network \citep{jaderberg2015}, $M\circ\Phi=f_{stn}(M,\Phi)$, and a single-stream encoder-decoder architecture with skip connections (similar to U-Net \citep{ronneberger2015}) is used. Two volumes, $M$ and $F$, are stacked as  the input of VoxelMorph. More details of VoxelMorph are described in \citep{balakrishnan2018, balakrishnan2019}.

\subsection{Dual-stream Architecture}
\begin{figure*}[htbp]
	\centering
	\includegraphics[width=16cm,height=4cm]{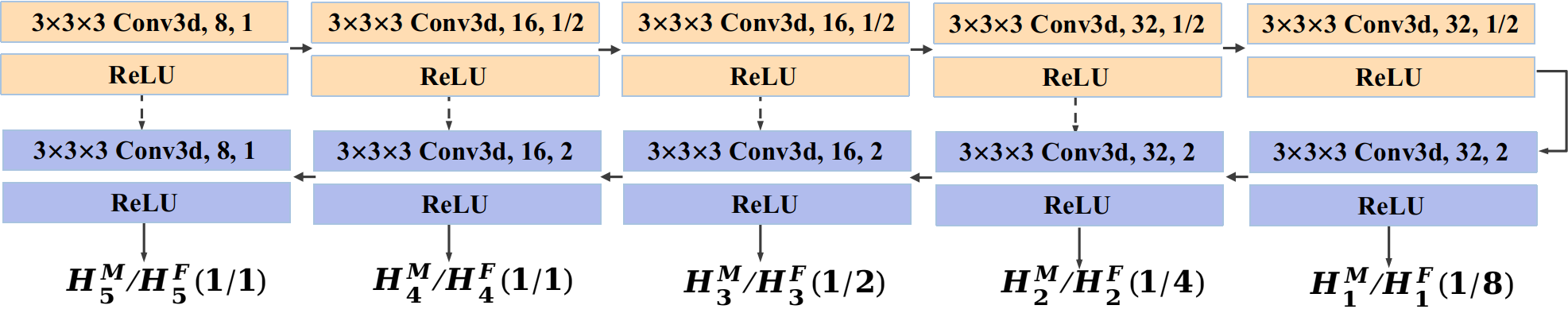}
	\caption{Backbone of the proposed Dual-PRNet$^{++}$. It consists of an encoder (yellow) and a decoder (blue), each of which has five convolutional blocks. The convolutional layers are indicated by the filter size, the number of output channels, and spatial resolution (w.r.t. the feature maps of previous layer). $H_{l}^M$ and $H_{l}^F$ denote the feature maps computed from the moving volume and the fixed volume separately, with various spatial resolutions (w.r.t. the input volumes) at different convolutional blocks. %Notice that the feature maps generated by the last encoding layer are  $H_0^M$, having an 1/16 spatial resolution of the input volumes.
	}  \label{fig:backbone}
\end{figure*}

Our Dual-PRNet$^{++}$ is built on the encoder-decoder architecture of VoxelMorph, but improves it by introducing a dual-stream design, as shown in Fig. \ref{fig:dual_prnet}.
Specifically, the backbone of Dual-PRNet$^{++}$ consists of a two-stream encoder-decoder with shared parameters.
%we remove the residual blocks and refine-unit in VoxResNet~\citep{chen2018}, and degrade the encoder as
We apply an encoder with the same architecture of U-Net~\citep{ronneberger2015}, which contains five convolutional blocks. Except for the first convolutional block, each block has a 3D down-sampling convolutional layer with a stride of 2, followed by a ReLU operation. Thus the encoder reduces the spatial resolution of the input volumes by a factor of 16 in total, as shown in Fig. \ref{fig:backbone}.
In the decoding stage, we apply skip connections to the corresponding convolutional maps in the encoding and decoding process. The lower-resolution convolutional maps (from decoding layers) are up-sampled and concatenated with the higher-resolution ones (from encoding layers), following by a 3$\times$3$\times$3 convolution layer and ReLU operation, as shown in Fig. \ref{fig:dual_prnet}. Finally, we obtain two feature pyramids with multi-resolution convolutional features computed from the moving volume and the fixed volume separately.

The proposed dual-stream design allows us to compute feature pyramids from two input volumes separately, and then predict deformable fields from the learned, stronger and more discriminative convolutional features, which is the key to improve the performance.
This is different from existing single-stream networks, such as \citep{balakrishnan2018, balakrishnan2019} and \citep{kuang2019}, which compute the convolutional features from two stacked volumes, and estimate the deformation fields using single-stream convolutional filters. Furthermore, our dual-stream architecture can compute two paired feature pyramids where layer-wise deformation fields can be estimated sequentially at multiple scales. This allows the model to generate a sequence of deformation fields by designing a new sequential pyramid registration method.

Notice that we modify the backbone applied in the original Dual-PRNet \citep{hu2019} by increasing the convolutional blocks from four to five, but reducing the number of channels from 32 at each layer to [8, 16, 16, 32, 32] for the five layers, and also removing the refine units in the original design to keep a lightweight and effective model. This results in a large reduction of the model parameters from 410K to 175K (which might also alleviate the potential overfitting), but maintains the similar performance. For example, it improves the Dice score from 0.631 to 0.653 on the Mindboggle101, but has a reduction with 0.767$\rightarrow$0.743 on the LPBA40. In this work, we will use the new backbone for Dual-PRNet$^{++}$ in all our experiments.

\subsection{Sequential Pyramid Registration}
VoxelMorph computes a single deformation field from the convolutional features at the last up-sampling layer in the decoding process,  which might make it difficult to handle multi-scale deformations precisely, which are often in case for different anatomical structures of the brain.  In this work, we propose a new pyramid registration by designing a set of pyramid registration (PR) modules, which are implemented sequentially at each decoding layer. This allows the model to predict multi-scale deformation fields with increasing resolutions, generating a sequence of pyramid deformation fields, as shown in Fig. \ref{fig:dual_prnet}.

%\subsubsection{Pyramid Registration (PR) Modules}
\textbf{PR module.} Each PR module estimates a deformation
field at each decoding layer. As input, the PR module uses a pair of convolutional features, together with a deformation field computed from the previous layer (except for the first decoding layer where the deformation field  is not available). As output, the PR module yields an estimated deformation field at a given resolution level, which is used in the next pyramid level.
The  PR module includes a sequence of operations with feature warping,
stacking, and convolution (as shown in Fig. \ref{fig:pr_module} (a)), which are implemented repeatedly over the decoding layers.

%This results in a sequence of deformation fields with increasing resolutions, starting from the lowest-resolution decoding layer to the highest-resolution one. Our network includes four decoding layers, and thus generates four deformation fields sequentially.
%\subsubsection{Sequential Operations}

\textbf{Sequential operations.} Specifically, the first deformation field ($\Phi_1$) is computed at the first decoding layer. We first stack the two convolutional features computed at the first decoding layer, and then apply a 3D convolution with size of 3$\times$3$\times$3 to estimate a deformation field.  The deformation field ($\Phi_1$) is 3D  maps with the same shape of the corresponding convolutional feature maps.  It is able to extract coarse-level context information, such as high-level anatomical structure of the brain, which is then encoded into the convolutional features computed at the next decoding layer via feature warping: (i) the current deformation field is up-sampled by using bilinear interpolation with a factor of 2, denoted as $ u\left(\Phi_1\right)$,  and (ii) then it is applied to warp the convolutional maps of the moving volume in the next layer, by using a grid sample operation, as shown in Fig. \ref{fig:pr_module} (a). Then the warped convolutional maps are stacked again with the corresponding convolutional features generated from the fixed volume, followed by a convolution operation to estimate a new deformation field.  This process is implemented repeatedly at each decoding layer, and can be formulated as,

\begin{equation}
 \Phi_l=C_l^{3\times3\times3}\left(H^M_l\circ u\left({\ \Phi}_{l-1}\right),\ H^F_l\right)
\end{equation}

%and is empirically set to 4
where $l=1,2,...,N$, indicates the number of decoding layers. $C_l^{3\times3\times3}$ denotes a 3D convolution at the $l$-th decoding layer.  The operator $\circ$ is the warping operation that maps the coordinates of $H^M_l$ to $H^F_l$ using $u\left({\ \Phi}_{l-1}\right)$, where $H^M_l$ and $H^F_l$ are the convolutional feature pyramids  computed from the moving volume and the fixed volume at the $l$-th decoding layer.

%Finally, the estimated deformation fields are warped sequentially and recurrently with up-sampling, to generate a final deformation field, which encodes meaningful multi-level context information with multi-scale deformations.

\subsection{PR$^{++}$ Modules}
\begin{figure*}[htbp]
	\centering
\subfloat[PR module] {\includegraphics[width=0.45\textwidth,height=5cm]{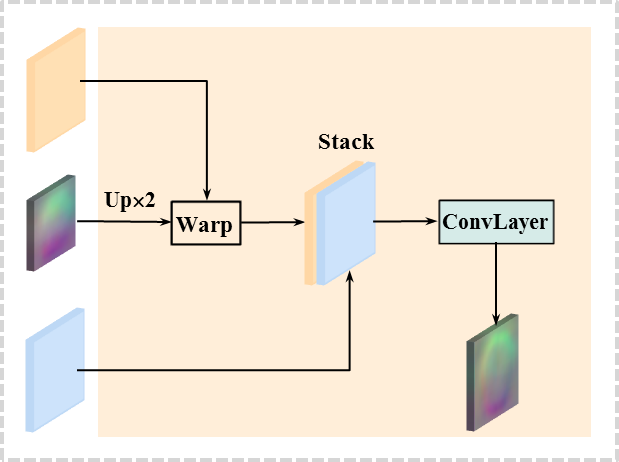}}
\subfloat[PR$^{++}$ module] {\includegraphics[width=0.5\textwidth,height=5cm]{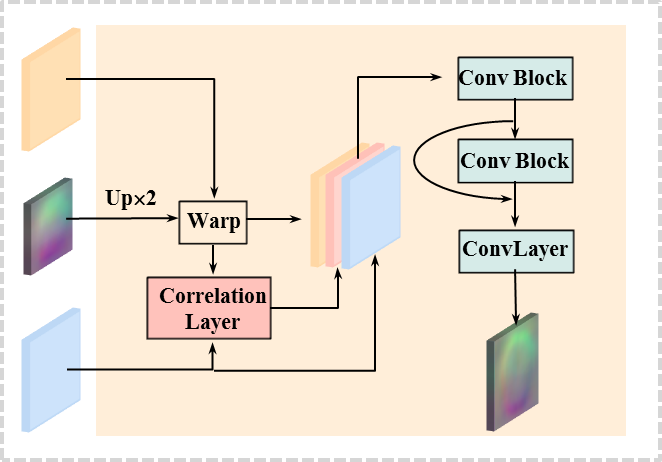}}
	\caption{The proposed (a) Pyramid Registration (PR) module, and (b) its extension: PR$^{++}$ module, which improves  the  PR  module  by  computing correlation features with further enhancement by  residual convolutions. } \label{fig:pr_module}
\end{figure*}

Sequential pyramid registration with a set of PR modules was originally introduced in our preliminary version \citep{hukang2019}. In this extension, we introduce PR$^{++}$ modules to enhance sequential pyramid registration. It improves the PR module by computing correlation features which are further enhanced by residual convolutions, as shown in Fig. \ref{fig:pr_module} (b). With respect to the PR module,  the PR$^{++}$ module includes two additional operations: 3D correlation and residual convolution, which are the key to enhance the learned features and in turn to boost the performance. Specifically, we design a 3D correlation layer to compute correlation features between the warped  features (from the moving volume) and the features from the fixed volume (see Fig. \ref{fig:pr_module} (b)). Then the correlation features, together with the two stacked features, are further processed by two convolution blocks with a residual connection to further enhance the representation.

%In this work, we design a pyramid registration module to process the warped and stacked convolutional features computed from previous stage. As shown in Fig. \ref{fig3}, the pyramid registration module consists of two key components: a 3D correlation layer and a feature fusion block, which enable the model to learn the strong correspondence between two volumes in the feature space.

\textbf{3D correlation layer}. In PR$^{++}$ module, a 3D correlation layer is designed to compute the local correlations between the two input volumes in the convolutional feature space. This allows us to aggregate the correlated features which are not directly explored in the original PR module, but can emphasise local details in deep representation.

%We first design a 3D correlation layer to measure the matching cost between the feature maps computed from the moving volume and the warped feature maps from the fixed volume.

Specifically, let $P^W_i$ and $P^F_j$ denote the central voxel of the 3D blocks (with size of $ (2k+1)^{3} $) sampled from the feature maps of the warped moving volume and the fixed volume. The correlation relationship between the two sampled 3D blocks can be computed as:

\begin{equation}
C\left(w_i,f_j\right)\ =\ \frac{1}{{(2k+1)}^3} \sum_{n_w, n_f \in [-k,k]^3}p_{i+n_w}^W \times p_{j+n_f}^F
\end{equation}

where $ n \in [-k,k]^3 $ means $ n $ iterates over a 3D neighborhood $ [-k,k]\times[-k,k]\times[-k,k] $ of $P^W_i$ or $P^F_j$. In our experiments, $k$ was empirically set to 1. Given a local 3D block on the feature maps of the (warped) moving volume, it is time-consuming to compute the dense correlations over all the 3D blocks sampled from the
feature maps of the fixed volume.
%
%Therefore, given a 3D block with $P^W_i$, we only compute the correlations within the corresponding neighborhood with a 3D region of $d \times d \times d $ of $P^F_j$ by striding over the two feature maps, which can be implemented as 3D convolutions.
Therefore, given a 3D block with $P^W_i$, we only compute the local correlations by sampling a set of $P^F_j$  within a 3D neighborhood of $d \times d \times d $, which can be implemented as 3D convolutions.
%which means $ P^F_j $ is limited to the neighbor block $ \Omega:[-k,k] \times [-k,k] \times [-k,k] $ of $ P^F_i $.
We use a stride $s_w = 1$ to densely sample $P^W_i$ from the warped feature maps, and set the correlation neighborhood with $d = 3$ on the corresponding fixed feature maps, where $ P^F_j $ is sampled with a stride of $s_f = 2$. Each sampled block has the same size of $ [-k,k]\times[-k,k]\times[-k,k] $ , and we compute direct correlations between two sampled blocks using E.q. (4).
This generates 3D correlation maps ($P^C$) with shape of $[2\times{FL}(d/s_f)+1]^3 \times(H/s_w)\times(W/s_w)\times(D/s_w)$,  where $[2\times{FL}(d/s_f)+1]^3 = 27$ is the number of channels. $FL$ indicates a $Floor$ computation.  The generated correlation maps have the same 3D shape as the feature maps of the moving and fixed volumes, which ensure that the three maps can be stacked together for further processing.

%By organizing the correlations $ C(w_i,f_j) $ in channels according to relative displacements, we obtain a $ [2(k/s_f)+1]^3 \times (H/s_w) \times (W/s_w) \times (D/s_w)$ correlation maps($ P^C $)  from a pair of feature maps with the size of $ H \times W \times D $.

\textbf{Convolutional enhancement.}
Our dual-stream architecture computes two separate feature pyramids  from two input volumes. However, the key to the registration task is to learn the strong anatomical correspondence between the two volumes in the feature space, which inspired us to design a new mechanism to further aggregate the computed pyramid features. The key function of the proposed PR$^{++}$ module is to provide a powerful approach for learning richer local details from the two features, which ensure more accurate estimations of the deformation fields at multiple levels. To enrich the learned features, the computed correlation maps are stacked with the two pyramid features at each decoding layers: the warped features from the moving volume and the pyramid features from the fixed volume, as shown in Fig. \ref{fig:pr_module} (b). The correlation maps have 27 channels over all decoding layers, while the number of channels of the two pyramid features varies over different layers: [8, 16, 16, 32, 32] for the five decoding layers in our experiments.

%which can be further enhanced  by sequential feature warping. However, the key to the registration task is to learn strong anatomical correspondence between the two volumes in the feature space, which inspired us to design a new mechanism to aggregate the computed pyramid features.

To this end, we apply two 3D convolution blocks for further processing the stacked features, as shown in Fig. \ref{fig:pr_module} (b).
Each convolution block consists of two 3$\times$3$\times$3 convolutional layers, followed by a ReLU operation.
The first convolution block reduces the channels of the stacked features considerably from [43, 59, 59, 91, 91] to [8, 16, 16, 32, 32] at the five decoding layers, which are consistent with the numbers of channels applied in the PR module for computational efficiency.
In addition, a residual connection is applied to the second convolutional block, in an effort to preserve more context information, and at the same time, to extract discriminative features between the two volumes. Finally, a convolution layer is used to estimate the deformation field. The new PR$^{++}$ module is applied to our dual-stream registration framework, resulting in the enhanced version Dual-PRNet$^{++}$, which boost the performance of the original Dual-PRNet, as demonstrated in our experiments.

\subsection{Final Deformation Field}
\begin{figure}[ht]
	\centering
	\includegraphics[scale=.31]{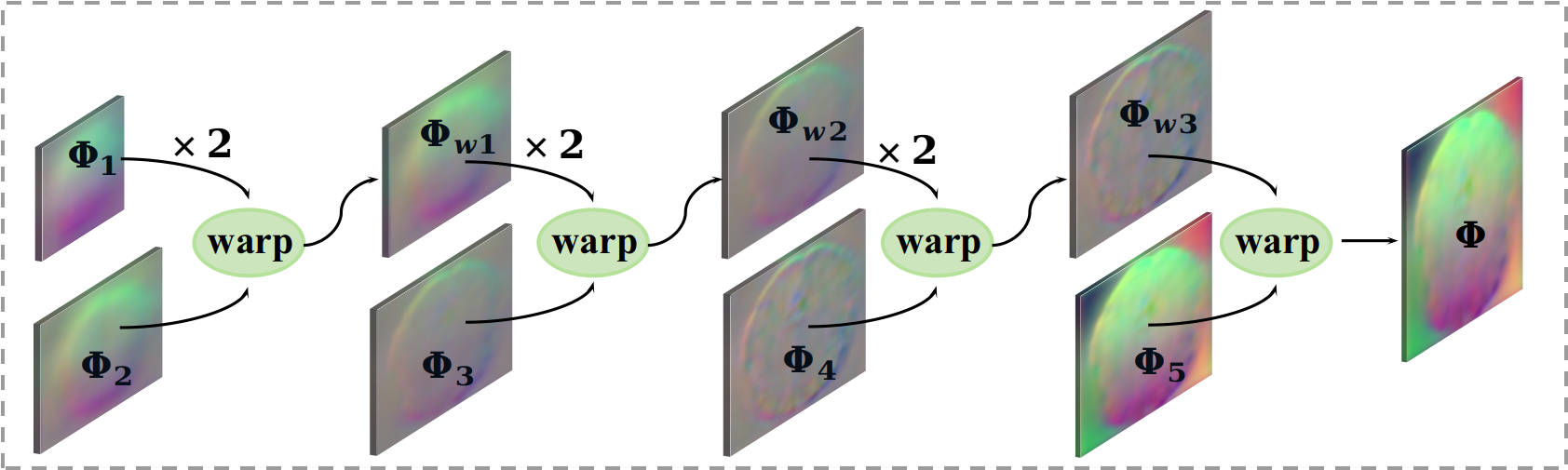}
	\caption{The final deformation field is computed by sequentially warping the current field with the previous one ($\times 2$ up-sampling).} \label{fig:final_DF}
\end{figure}

The proposed Dual-PRNet$^{++}$ generates five sequential deformation fields with increasing resolutions, indicated as [$\Phi_1$, $\Phi_2$, $\Phi_3$, $\Phi_4$, $\Phi_5$].
To compute the final deformation field, an estimated deformation field is up-sampled by a factor of 2, and then is warped by the following deformation field being estimated. Such up-sampling and warping operations are implemented repeatedly and sequentially to generate the final deformation field (as shown in Fig. \ref{fig:final_DF}), which encodes rich multi-level context information with multi-scale deformations.
This allows the model to propagate strong context information over hierarchical decoding layers, where the estimated deformation fields are refined gradually in a coarse-to-fine manner, and thus aggregate both high-level context information and low-level detailed features.
The high-level context information equips our model with the ability to work with large-scale deformations, while the fine-scale features allows it to model detailed anatomical structure information. We integrate PR modules or PR$^{++}$ modules into our dual-stream architecture, resulting in an end-to-end trainable model. By simply following VoxelMorph~\citep{balakrishnan2018, balakrishnan2019}, a negative local cross correlation (NLCC) is applied as the loss function, coupled with  a smooth regularization, e.g., a diffusion regularizer which computes approximate spatial gradients using differences between neighboring voxels, as detailed in~\citep{balakrishnan2018}.

\section{Joint Segmentation and Registration}

\begin{figure*}[tb]
	\centering
	\includegraphics[width=15cm,height=5.5cm]{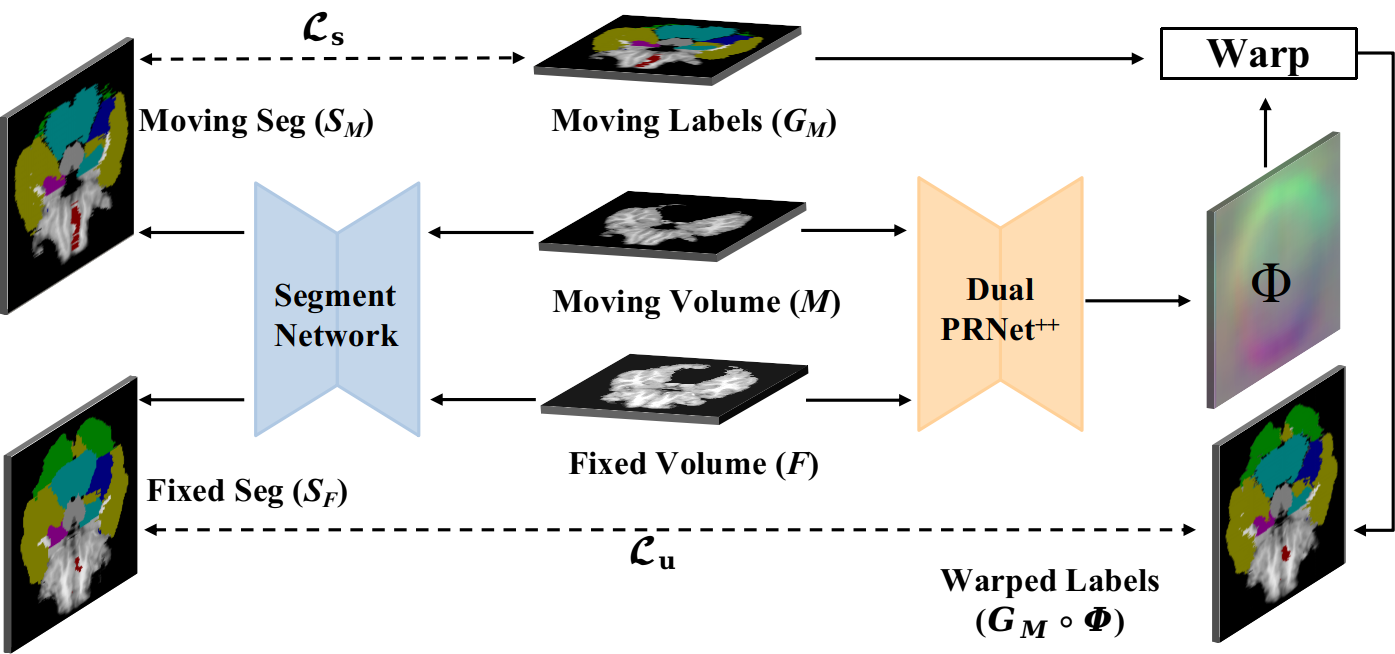}
	\vspace{-3mm}
	\caption{Joint segmentation and registration framework, where the proposed Dual-PRNet$^{++}$ is used as a joint registration network, and is trained jointly with a 3D segmentation network. Dual-PRNet$^{++}$ is applied to warp the moving labels, which are then used as supervision of the corresponding fixed volume.} \label{fig:joint}
\end{figure*}

Recent segmentation methods using deep learning technologies often require massive manually annotated data, which is labor-intensive and expensive, particularly for 3D medical images.
It is appealing to develop unsupervised or weakly supervised methods for accurate segmentation on 3D medical images. The proposed Dual-PRNet or Dual-PRNet$^{++}$ is able to transfer a  moving volume to a fixed volume, which inspired us to adopt such ability to roughly map the available segmentation labels from a source domain to a target domain where the annotations are not provided. This enables us to train a segmentation network on the target MRI domain by using the transferred anatomical labels, without any manual annotation.

In this work, we integrate the registration network into a segmentation network to form a unified framework, as shown in  Fig. \ref{fig:joint}. The framework is related to that of \citep{xu2019} where DeepAtlas was developed  to learn the two tasks simultaneously by using Voxelmorph as the registration network. We extend the DeepAtlas approach by using the proposed Dual-PRNet$^{++}$ as the registration network.
%which is able to provide more accurate warped anatomical labels for the unlabelled volumes, resulting in considerable performance improvements on the segmentation task. 
Notice that the  pre-trained registration network is fixed during the training of segmentation network. 

%Furthermore, a more efficient solution is to further integrate the registration network into a segmentation network to learn both segmentation and registration jointly in a unified framework, as shown in Fig. \ref{fig:joint}. Our joint framework is related to that of \citep{xu2019} where DeepAtlas was developed  to learn the two tasks simultaneously by using Voxelmorph as the registration network. In this work, we extend the DeepAtlas approach by using the proposed Dual-PRNet$^{++}$ as the registration network, which is able to provide more accurate warped anatomical labels for the unlabelled volumes, resulting in considerable performance improvements on the joint segmentation.

%Without any annotation, the unsupervised registration network can predict the deformation field, which can be used to align the anatomical label to the target volume and provide large amount of coarse labeled data effectively. This inspires Xu \emph{et al.}\citep{xu2019} propose a framework called DeepAtlas to jointly learn the segmentation and registration. In this paper, we replace the Voxelmorph on DeepAtlas with our Dual-PRNet$_{++}$ and integrated it into the segmentation network, which achieves the performance improvement by providing more accurate warped label for segmentation.

Details of  the unified  framework of segmentation and registration  are presented in Fig. \ref{fig:joint}. Given a pair of moving ($M$) and fixed volumes ($F$), a registration network is adopted to estimate  a deformation field ($\Phi$), which is then used to warp the available  segmentation labels from the source volume to the unlabelled (target) one, e.g., from the moving volume to the fixed one. Taking the moving and fixed volumes as input, the segmentation network generates two segmentation maps, denoted as $S_M$ and $S_F$. The source volumes which have ground-truth labels ($ G $) are utilized to train the segmentation network in a regular supervised manner, while the unlabeled volumes can be used to train the same segmentation network,  by leveraging the generated labels warped from the corresponding volumes having labels. Specifically, when the moving volumes are labeled and the fixed volumes are unlabeled, the segmentation loss ($\mathcal{L}_{seg}$) for the unified framework can be computed as:

\begin{equation}
\mathcal{L}_{seg}=\lambda_M\mathcal{L}_s\left(S_M, G_M\right)+\lambda_F\mathcal{L}_u\left(S_F, G_M\circ\Phi\right) \label{eq:joint1}
\end{equation}
where $\mathcal{L}_s$ and $\mathcal{L}_u$ are the segmentation losses computed from the moving volumes (with labels) and the fixed volumes (without labels) respectively. In this paper, we adopt a Dice loss for the segmentation task by following DeepAtlas \citep{xu2019}. $\lambda_M$ and $\lambda_F$ are the weights that balance the impact of labelled and unlabeled data. Conversely, the two segmentation losses can be computed reversely when we use the  moving volumes as unlabeled and the fixed ones as labeled:
\begin{equation}
\mathcal{L}_{seg}=\lambda_M\mathcal{L}_s\left(S_F, G_F\right) +\lambda_F\mathcal{L}_u\left(S_M\circ\Phi, G_F\right) \label{eq:joint2}
\end{equation}

\begin{figure*}[tb]
\centering
\subfloat[Moving] {\includegraphics[width=2.2cm, height=4cm]{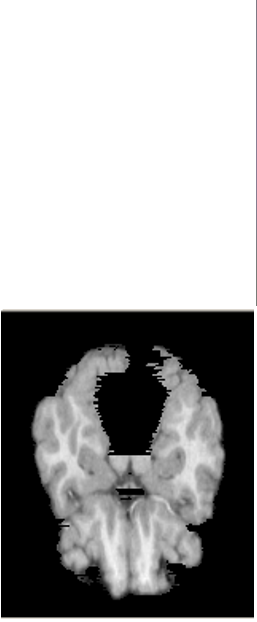}}
\subfloat[The estimated fields with the corresponding warped volumes] {\includegraphics[width=11cm, height=4cm]{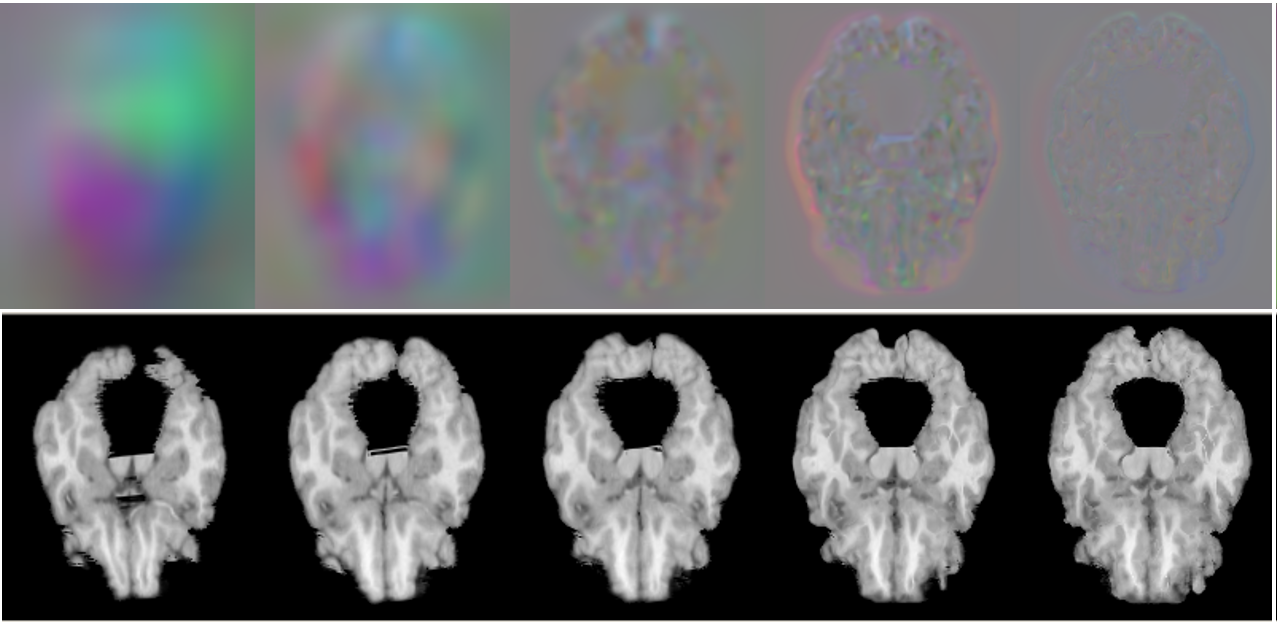}}
\subfloat[Final field] {\includegraphics[width=2.2cm, height=4cm]{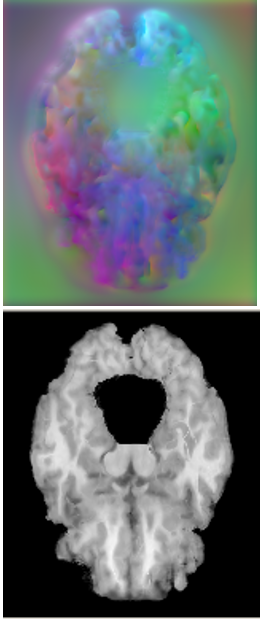}}
\subfloat[Fixed] {\includegraphics[width=2.2cm, height=4cm]{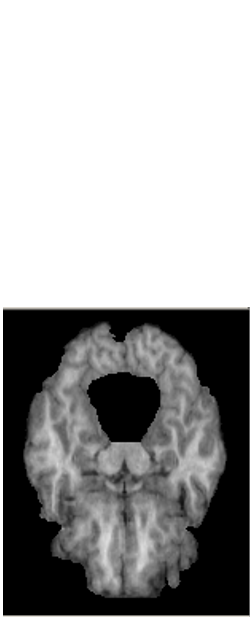}}

%	\vspace{-4mm}
\caption{Visualization of the estimated deformation fields with the corresponding warped volumes: (a) the moving volume, (b) the estimated deformation fields with increasing resolutions (top: left$\rightarrow$right), with the corresponding warped volumes (bottom: left$\rightarrow$right). Sequential warping is implemented with all the preceding fields, and thus the last warped volume is exactly the same as the volume warped by the final field. (c) the final deformation field with the warped volume, and (d) the fixed volume.} \label{fig:visualization}
\end{figure*}

\section{Experimental Results and Comparisons}
\textbf{Datasets.}
The proposed Dual-PRNet and Dual-PRNet$^{++}$ are evaluated on 3D brain MRI registration on two public datasets, LPBA40~\citep{shattuck2008} and Mindboggle101~\citep{klein2012}.
The LPBA40~\citep{shattuck2008} contains 40 T1-weighted MR images, each of which was annotated with 56 subcortical ROIs.
The Mindboggle101~\citep{klein2012} has 101 T1-weighted MR images, which were annotated with 25 cortical regions or 31 cortical regions, and can be used to evaluate registration results with more fine and detailed structure of the brain.

\textbf{Experimental settings.}  Our experiments on unsupervised  3D  brain  MRI  registration were conducted by following~\citep{kuang2019}. Specifically, on the LPBA40, we train our models on 30 subjects, generating 30$\times$29 volume pairs, and test on the remaining 10 subjects.
  We follow ~\citep{kuang2019} with the provided code\footnote{https://github.com/dykuang/Medical-image-registration.}, and merge 56 labels into 7 regions specified as: Frontal Lobe, Parietal Lobe, Occipital Lobe, Temporal Lobe, Cingulate Lobe, Putamen, and Hippocampus, which were defined by the major clinical structures of the brain (e.g., each cortical lobe, plus three more regions).  Then we center-crop the
volumes into a size of $ 160\times192\times160 $. 
 On the Mindboggle101, we adopt the 25 cortical regions in our experiments on the registration task, and further merge the 25 cortical regions into five large regions corresponding to five anatomical structures of the brain: Frontal lobe, Parietal lobe, Occipital lobe, Temporal lobe, and Cingulate lobe, again by following the implementation details of \cite{kuang2019}.  The data was divided into 42 subjects (with 1722 pairs) for training, and 20 subjects with 380 pairs for testing.  All volumes were cropped with size of $ 160\times192\times160 $.

For joint segmentation and registration, we conducted experiments on the Mindboggle101 with 31 cortical regions by following \citep{xu2019}. $\lambda_M$ and $\lambda_F$ in E.q. \ref{eq:joint1} or \ref{eq:joint2} are set to 1 in our experiments.

The proposed Dual-PRNet and Dual-PRNet$^{++}$ were implemented in Pytorch and trained on 4 Titan Xp GPUs.  Batch size is set to 4, due to the limitation of GPU memory. We adopt Adam optimization with a learning rate of 1e-4. The results of VoxelMorph were produced by running the codes provided by the original authors.
%Limited by the memory of GPU, for correlation layer, we set $ k=1, d=3, s_m=1, s_w=2$ empirically and produce multi-scale correlation maps with 27 channels.

\textbf{Measurements.} We adopted the Dice score,  Average Symmetric
Surface Distance (ASD  in mm), and Symmetric Hausdorff Distance(HD  in mm),   by following \citep{de2019deep},
%and folding fractions of Jacobian determinant on deformation field \citep{Ashburner2007A} 
as evaluation metrics. The Dice score measures the degree of overlap at the voxel level. 
\begin{equation}
\mathcal Dice= \frac{2\left| L_W\cap L_F \right|}{\left| L_W \right| + \left| LF \right| }
\label{eq:dice}
\end{equation}
where $L_W$ and $L_F$ denote the labels of warped volume and fixed volume. 
ASD and HD calculate a surface distance between the moved label and fixed label, which are sensitive to outliers of registration results. Given $\mathcal{R}_W$ and $ \mathcal{R}_F $ as the surface point sets of the warped label and fixed label, we can computed the ASD as follows:
\begin{equation}
\mathcal ASD = \frac{\sum_{x\in \mathcal{R}_W}D(x,  \mathcal{R}_F) + \sum_{y\in \mathcal{R}_F}D(y, \mathcal{R}_W) }{\left|  \mathcal{R}_W \right| + \left| \mathcal{R}_F \right| }
\label{eq:asd}
\end{equation}
where $ D(x, \mathcal{R}_F) $ denote a minimal distance of one point $ x $ to another point in $\mathcal{R}_F$. Additionally, we adopt the definition of HD as: 
\begin{equation}
\mathcal HD = \max\left\{ D_h(\mathcal{R}_W, \mathcal{R}_F), D_h(\mathcal{R}_F, \mathcal{R}_W) \right\}
\label{eq:hd}
\end{equation}
where 
\begin{equation}
\mathcal D_h(\mathcal{R}_W, \mathcal{R}_F) = \max_{x\in\mathcal{R}_W} \min_{y\in\mathcal{R}_F}D\left( x, y \right)
\label{eq:Dh}
\end{equation}

In addition, to measure the smoothness of the estimated deformation field, we further compute folding fractions of Jacobian determinant on the field \citep{Ashburner2007A}. 
The Jacobian determinate $\left| J_{\Phi} \right| $ on a deformation field indicates the relative changes in a local area. Specifically, $\left| J_{\Phi}(p) \right| \leq 0 $ means the folding has occurred around the location $ p $ of $\Phi$ , which means $\Phi$ is non-smooth and not physically realistic. Therefore, we adopt the fraction of folding on $ \left| J_{\Phi} \right| $ to evaluate the regularity of deformation field.

\subsection{Comparisons with the State-of-the-Art Approaches}
\begin{figure}[ht]
	\centering
	\includegraphics[scale=0.45]{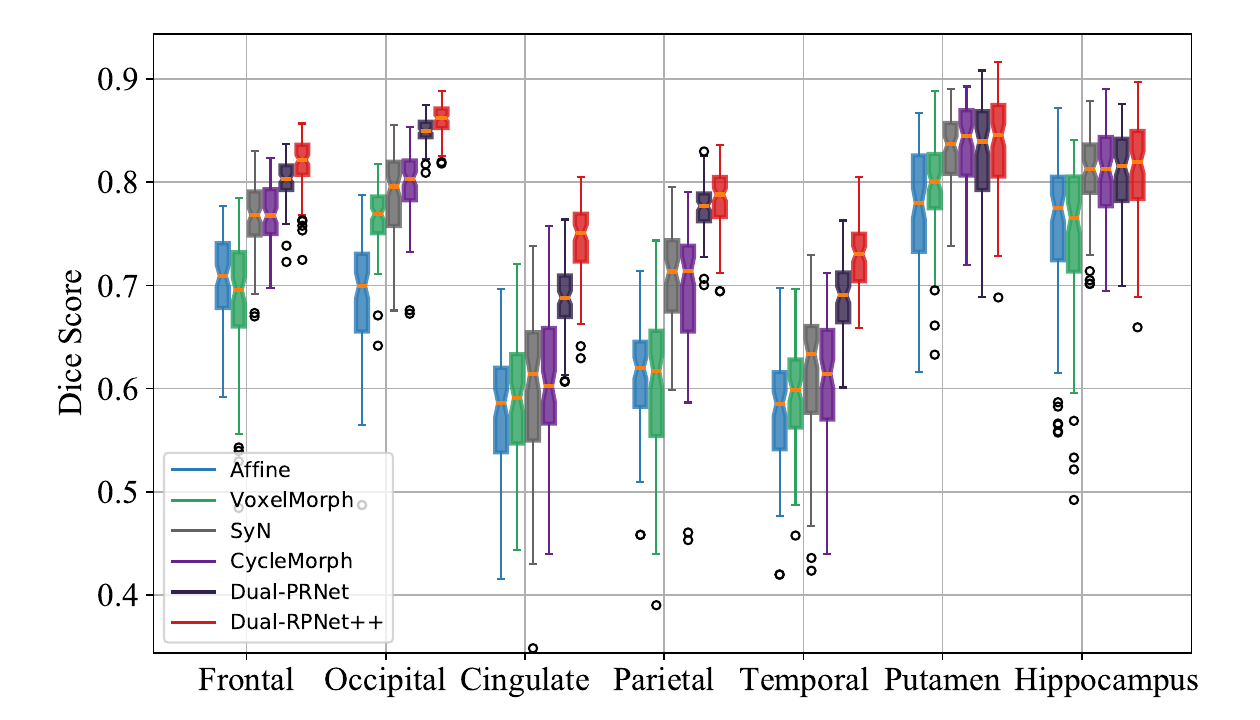}
	\caption{Dice scores of different methods on LPBA40 (7 regions). The average scores are: 0.669 (Affine), 0.683 (VoxelMorph), 0.731 (SyN), 0.733 (CycleMorph), 0.778 (Dual-PRNet), and 0.798 (Dual-PRNet$ ^{++}$).}\label{fig:LPBA40}
\end{figure}

We compare our Dual-PRNet and Dual-PRNet$^{++}$ with a number of recent approaches: affine registration, SyN~\citep{shattuck2008}, VoxelMorph~\citep{balakrishnan2018, balakrishnan2019}, FAIM \citep{kuang2019}, PMRNet \citep{liu2019},  LapIRN \citep{mok2020large},   Contrastive Registration (CReg) \citep{Lihao2020} and CycleMorph \citep{KIM2021102036}  on both LPBA40 and Mindboggle101 datasets.  As discussed in Section 2, VoxelMorph estimates a single deformation field with a single-stream network. FAIM extends VoxelMorph by designing a new penalty loss on negative Jacobian determinants. PMRNet computes average multi-resolution deformation fields, which are obtained from a dual-stream encoder, as the final deformation field. CReg estimates a single transformation field from feature pyramids by using a contrastive loss and a single-stream decoder.  Two registration networks were designed in CycleMorph \citep{KIM2021102036}, with a cycle consistency, which takes inverse order volumes as inputs.  We implemented the affine registration and SyN by using ANTs~\citep{avants2011}.  For VoxelMorph and FAIM, we used the codes and models provided by the original authors. For CycleMorph, we utilized the released code, and trained the models on the datasets used in our experiment, by using the same experimental settings as \citep{KIM2021102036}.
\begin{table}[!t]
		\caption{ The results of different methods on LPBA40, in the terms of average Dice Score (Avg Dice), Symmetric Hausdorff Distance (HD in mm), and Average Symmetric Surface Distance (ASD in mm).
}\label{tab:lpba_hd}
		%\vspace{-3mm}
		\centering
		\begin{tabular}{l|c|c|c}
			\hline
			   &Avg Dice$\uparrow$  &  HD$\downarrow$  &  ASD$\downarrow$\\
			\hline
			Affine&0.669 & 13.283 & 2.469 	\\
			VoxelMorph&0.683 & 14.575  & 2.238 	\\
			FAIM &0.664 & 12.935  & 1.790 \\
			CycleMorph&0.733 & 12.961  & 1.886 	\\
			 LapIRN&  0.796 &   \textbf{11.824} &  \textbf{1.504} \\\hline
			Dual-PRNet&0.778& 13.549  & 2.096 	 \\
			Dual-PRNet$^{++}$&\textbf{0.798} & 12.983  & 1.724 \\ \hline
					\hline
		\end{tabular}
\end{table}

\begin{table*}[!t]
		\caption{The results of different methods on Mindboggle101, in the terms of average Dice Score (Avg Dice), Symmetric Hausdorff Distance (HD in mm), and Average Symmetric Surface Distance (ASD in mm).
%Dual-PRNet$ ^{++} $ (CD) indicates cross-dataset (CD) evaluation.
}\label{tab:mindboggle101}
		%\vspace{-3mm}
		\centering
		\begin{tabular}{l|c|c|c|c|c||c|c|c}
			\hline
			Region  &  Frontal  &  Parietal  &  Occipital  &  Temporal  &	Cingulate  &  Avg Dice$\uparrow$  &  HD$\downarrow$  &  ASD$\downarrow$ \\\hline
			Affine&	0.455&	0.406&	0.354&	0.469&	0.450&	0.427 &  16.27 & 1.433 \\
			SyN& 	0.558&	0.496&	0.446&	0.578&	0.549&	0.525 & --  & -- \\
			VoxelMorph&	0.532&	0.459&	0.480&	0.585&	0.499&	0.511& 16.977  & 1.261 \\
			FAIM&	0.572&	0.551&	0.537&	0.469&	0.508&	0.527& 16.544  & 1.018 \\
			PMRNet&	0.579&	0.559&	0.430&	0.544&	0.546&	0.532& --  & -- \\
			 LapIRN &  0.543&  0.634&  0.477& 0.632&  0.627&  0.583& \textbf{15.920} &   1.069 \\
			CReg&	0.644&	0.620&	0.537&	0.703&	0.640&	0.629& --  & -- \\
			CycleMorph& 0.695	 & 0.612  & 0.526	& 0.683	& 0.628 & 0.629&16.255  & 1.009	 \\\hline
			Dual-PRNet&0.602 &0.690&0.550&	0.695&	0.618&	0.631& 16.826  & 1.395 \\
			Dual-PRNet$^{++}$&\textbf{0.735} &\textbf{0.810}&	\textbf{0.667}&	\textbf{0.802}&	\textbf{0.724}&	\textbf{0.748} & 16.080  & \textbf{0.849} \\ \hline
%			Dual-PRNet$^{++}$(CD) & 0.727 & 0.803 & 0.660 & 0.795 &	0.711 &	0.739\\
			
			\hline
		\end{tabular}
\end{table*}

\textbf{Results and comparisons.} On the LPBA40 dataset, as shown in Fig. \ref{fig:LPBA40}  and Table \ref{tab:lpba_hd} , Dual-PRNet improves the average Dice score to 0.778,  and outperforms SyN, VoxelMorph, and CycleMoprh considerably. With the enhanced PR$^{++}$ module, Dual-PRNet$^{++}$ can further increase the average Dice score by 2.0\% (Dice 0.798), and achieves the best performance in the term of average Dice score.
However, Dual-PRNet$^{++}$ is outperformed by LapIR in the terms of ASD and HD.  With the penalty on both the size and nonsmoothness of the deformation field, LapIR is able to obtain the lowest HD and folding fraction on the determinate of the Jacobian.

% Furthermore, Dual-PRNet$^{++}$ achieves the best result on ASD, and a competitive performance with FAIM and CycleMorph in the term of HD. With the penalty both on the size and nonsmoothness of the deformation field, FAIM obtain the lowest HD and folding fraction on the determinate of the Jacobian.

The results on the Mindboggle101 are shown in Table \ref{tab:mindboggle101}, where our Dual-PRNet$^{++}$ consistently outperforms the other methods on Dice score, and achieves the best performance on all five regions. \textit{It reaches a high average Dice score of 0.748, surpassing the closest one - 0.629 of CReg and CycleMorph, by a large margin. Notice that the new PR$^{++}$ modules lead to a large improvement of $0.631\rightarrow 0.748$ over the original Dual-PRNet, demonstrating its ability to learn detailed brain structure}.
 Furthermore, Dual-PRNet$^{++}$ achieves an ASD of 0.849, which is the best result among all methods, and a comparable HD with LapIRN. This indicates that our method is able to generate less outliers when performing registration.
Notice that our methods can achieve excellent Dice scores by simply using CC loss and smooth loss, but did not explored an additional penalty on the deformation fields as did by LapIRN, which would reduce the HD and ASD.

In addition, we also computed folding fractions of Jacobian  determinant on deformation fields which measure the smoothness of the deformation fields. Our Dual-PRNet++ obtained a folding faction of 1.725 on the Mindboggle101, outperforming VoxelMorph with 2.274, while FAIM and CycleMorph achieved higher performance of 0.983 and 1.142 respectively. Notice that both FAIM and CycleMorph have a regularity loss designed specifically to encourage the smoothness of the estimated deformation fields in an explicit manner. For example, negative Jacobian determinants were directly measured as a loss in FAIM, while CycleMorph computes a regularization function and a cycle constraint loss to achieve it.

\subsection{Ablation Study}
We provide ablation studies to further verify the efficiency of individual technical components developed in our Dual-PRNet$^{++}$. We assessed the benefit of the dual-stream design, sequential pyramid registration with PR modules, and the improved PR$^{++}$ modules. Results from these ablation experiments on the Mindboggle101 are presented in Table \ref{tab:ablation}. To compare our dual-stream architecture with the single-stream design in VoxelMorph, we apply the dual-stream architecture for estimating a single deformation field as VoxelMorph, which achieved an average Dice score of 0.582 on the Mindboggle101 and 0.767 on the LPBA40, improving the single-stream counterpart by +7.1\% and  +8.4\%, respectively.
By integrating our sequential pyramid registration with PR modules, the results can be further increased, with  +4.9\% and +1.1\%  further improvements on the Mindboggle101 and LPBA40.
To further enhance the local details in the learned deep features,
we develop the PR$^{++}$ modules which aggregate richer local details by explicitly computing the local correlation features, with residual convolutions for further enhancement. This results in a large further improvement on the Mindboggle101: 0.631$\rightarrow$0.748 on Dice score, as shown in Table \ref{tab:ablation}.
 Particularly, the residual convolutions and 3D correlation have independent improvements of 0.631$\rightarrow$0.694 and 0.694$\rightarrow$0.748 respectively, making comparable contribution in our design. 
%In addition, the PR$^{++}$ module also yielded $+2\%$ improvement on the LPBA40 as indicated in Fig. \ref{fig:LPBA40}.
%
%

%or local correlation features can naturally make more contribution.

%Dual-PRNet$^{++}$, we investigate the power of correlation layer and feature fusion block. Table \ref{tab2} shows that, the feature fusion block improve the performance with ** based on the Dual-PRNet.

\begin{table}[!t]
	\caption{Ablation study on different components of Dual-PRNet $^{++}$ on Mindboggle101 (5 regions) and LPBA40, with average Dice scores reported.}\label{tab:ablation}
	%\hspace{-3mm}
	\centering
	\begin{tabular}{c|c|c|c|c|c}
		\hline
		Dual-stream & PR  &  Res. & Cor.  & Mind101 & LPBA40\\
		\hline
	    $\times$ & $\times$& $\times$& $\times$& 0.511& 0.683\\
		\checkmark & $\times$& $\times$& $\times$& 0.582& 0.767\\
		\checkmark &\checkmark& $\times$ & $\times$ & 0.631 & 0.778 \\
		 \checkmark &  \checkmark&  \checkmark &  $\times$ &  0.694 &  0.785 \\
		\checkmark &\checkmark& \checkmark & \checkmark &	0.748 & 0.798 \\
		\hline
	\end{tabular}
\end{table}

\subsection{Results on Joint Segmentation and Registration}
\begin{table}[!t]
	\caption{Performance of joint segmentation and registration on Mindboggle101 (31 regions).}\label{tab:joint}
	%\vspace{-3mm}
	\centering
	\begin{tabular}{l|c|c|c}
		\hline
		Model  &  N=1  &  N=21  &  N=65 \\
		\hline
		Segmentation &	-- --&	73.48&	81.31\\
		 DeepAtlas + VoxelMorph 	& 	61.19&	75.63&	-- --\\
		 DeepAtlas + Dual-PRNet$^{++}$  & \textbf{66.86}	&\textbf{78.04}&	-- --\\
		\hline
	\end{tabular}
\end{table}

\begin{figure*}[tb]
	\centering
	\includegraphics[width=0.95\textwidth, height=3.8cm]{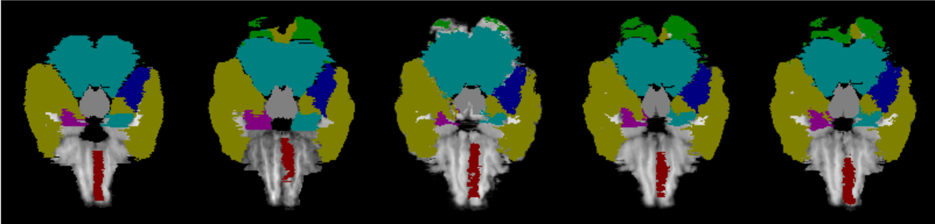}
	\includegraphics[width=0.95\textwidth, height=3.8cm]{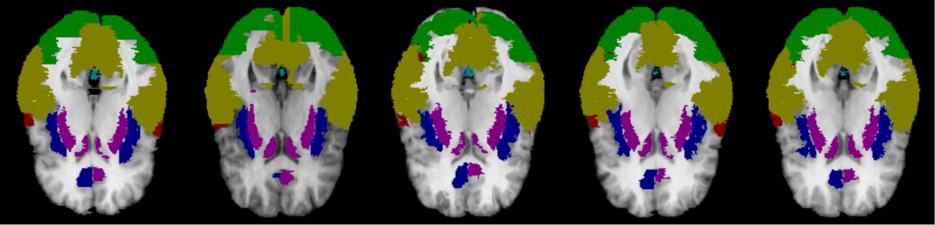}
	\includegraphics[width=0.95\textwidth, height=3.8cm]{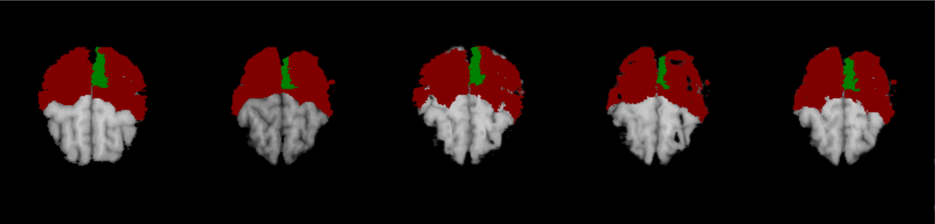}
	%\vspace{-4mm}
	\caption{Registration results on large spatial displacements. From left to right: the moving image, the fixed image, results of VoxelMorph, Dual-PRNet and Dual-PRNet$^{++}$.} \label{fig:displacements}
\end{figure*}

We further evaluate the performance of joint segmentation and registration framework by using the proposed Dual-PRNet$^{++}$, which can be integrated into a 3D segmentation network. To make a fair comparison, we replace  Voxelmorph with our Dual-PRNet$^{++}$ as the registration network in the joint framework of DeepAtlas \citep{xu2019}. Our Dual-PRNet$^{++}$ estimates a deformation field (the final one as described in Section 2.5) which is then used to warp the available segmentation labels from the source volume to the target one,  to guide the learning.  Notice that in this implementation, we only optimize the segmentation network by fixing the registration one (also referred as semi-DA in \cite{xu2019}), due to limited GPU memory. Higher performance can be expected with a fully joint learning of the two tasks. 

%, which is guided by the warped label or segmentation produced from registration network. In this form of data augmentation, limited labeled data can be increased significantly and even worked on one-shot situation.

%For fair comparison, we keep the same segmentation network and replace the Voxelmorph with Dual-PRNet$^{++}$ as registration network.
Experiments were conducted on the Mindboggle101.  By following DeepAtlas \citep{xu2019}, we use 31 labeled regions in the experiments which are different from the 25 regions used in previous registration experiments.  Again, with same experimental settings as \citep{xu2019},  the joint networks are trained with $N$ labeled volumes and the remained $65-N$  volumes unlabeled. It is a fully supervised learning when $N = 65$, which is the total number of the volumes in the dataset.  We use $N=21$ in our experiments by following DeepAtlas, and results are compared in Table. \ref{tab:joint}.

In the case of  $N=21$,  DeepAtlas with our Dual-PRNet$^{++}$  obtains an average Dice score of 78.04\%, clearly  outperforming the pure segmentation network (73.48\%)  which is only trained on 21 labelled volumes. This demonstrates that the joint registration network is greatly helpful to improve the performance of segmentation network, by leveraging the additional unlabelled volumes.
Furthermore, as the joint registration network,
our Dual-PRNet$^{++}$ can provide more accurate warped anatomical labels than  Voxelmorph used by DeepAtlas \citep{xu2019}, resulting in large performance improvements on the joint framework, e.g., $61.19\%\rightarrow 66.86\%$  in one-shot learning ($N=1$), and $75.63\%\rightarrow 78.04\%$ when $N=21$. Notice that our result (78.04\%) is also closed to the result (81.31\%) of fully supervised learning where all labelled volumes ($65$ in total) are used for training. The results suggest that our Dual-PRNet$^{++}$ can work more effectively in the joint segmentation and registration framework, and is helpful to training the segmentation network with limited data and labels provided.

%shows that our joint networks can achieve a higher accuracy by providing more accurate warped labels.

\section{Discussion}
In this section, we further study the robustness of our methods in the cases of large displacements and large splice spaces. Then we analyze the performance on  detailed structure  and cross-dataset learning, with a discussion on the limitation of our methods, which might accumulate interpolation artifacts.

\begin{figure*}[ht]
	\centering
	\includegraphics[width=0.95\textwidth]{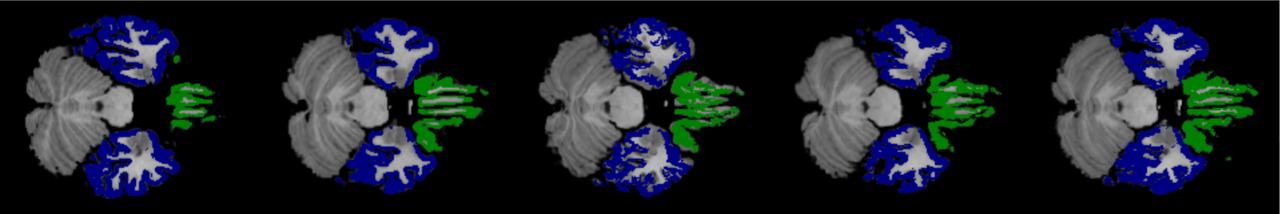}
	\includegraphics[width=0.95\textwidth]{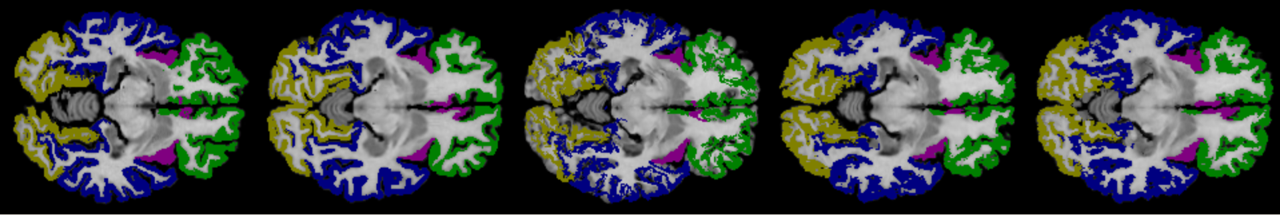}
	\includegraphics[width=0.95\textwidth]{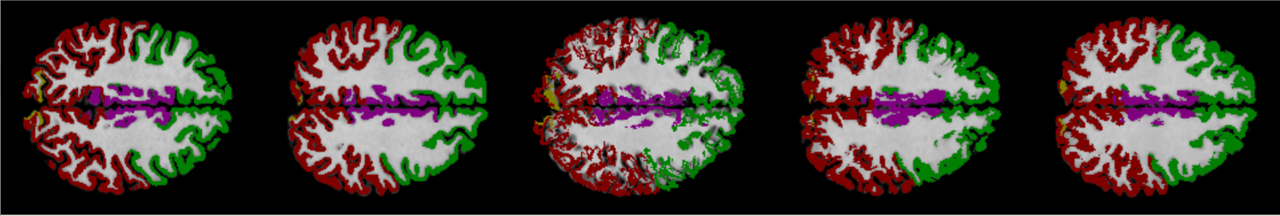}
	\vspace{-2mm}
	\caption{Registration results on the Mindboggle101. From left to right: the moving image, the fixed image, results of VoxelMorph, Dual-PRNet and Dual-PRNet$^{++}$.} \label{fig:Mindboggle101}
\end{figure*}

\subsection{Robustness}
\textbf{On large displacement.} We first visualize the generated multi-resolution deformation fields in Fig.~\ref{fig:visualization} (top). As can be found, a deformation field generated from a lower-resolution layer contains coarse and high-level context information, which is able to encode the high-level semantic information by warping a volume at a larger scale. Conversely, the deformation field estimated from a higher-resolution layer can capture more detailed features.
Fig.~\ref{fig:visualization} (bottom) shows the warped images by using the corresponding deformation fields presented. The sequential deformation fields are refined gradually to generate the final field, which can warp the moving image toward the fixed one more accurately, by aggregating more detailed structural information from the preceding fields via sequential warping, as shown in Fig. \ref{fig:visualization} (c).
%
%The warped images are refined gradually toward the fixed image, by aggregating more detailed structural information.
%

%\textbf{Results on large spatial displacement.}
We investigate the capability of our methods for handling large spatial displacements, and compare our registration results against that of VoxelMorph in Fig.~\ref{fig:displacements}.
Our Dual-PRNet and Dual-PRNet$^{++}$ can align the image more accurately than VoxelMorph, especially on the regions containing large spatial displacements, as indicated in green or red regions. In addition, as can be found, Dual-PRNet$^{++}$ has an improvement over the original Dual-PRNet, by using the enhanced PR$^{++}$ modules. The design of  3D correlations with more convolutional layers in the PR++ modules can enlarge the receptive fields, which in turn further enhances the ability to handle large displacements.

To further verify the ability of our Dual-PRNet$^{++}$ to handle large displacements quantitatively, we assume that the large displacements more likely happen on the regions where an affine registration can not perform well, such as the ``Occipital" and ``Parietal" regions on Mindboggle101, which have low Dice scores of 0.354 and 0.406 respectively.  Our Dual-PRNet$^{++}$ can have large \textit{relative} improvements of  88\%-100\% in these two regions (where VoxelMorph only has 13\%-36\%  \textit{relative} improvements),
compared to 60\%-71\%  \textit{relative} improvements on the regions where the  affine registration achieves a higher Dice score over 0.450.
%compared to the 13\%-36\%  \textit{relative} improvements by VoxelMorph. 
%
On the LPBA40, our  Dual-PRNet$^{++}$ has  \textit{relative} improvements of 26\%-29\% on the regions of ``Cingulate" and ``Temporal" which have low Dice scores of 0.576 and 0.578 by the affine registration, while only achieving about 8\% \textit{relative} improvements on the regions of ``Hippocampus" and ``Putamen" where the affine registration performs better with 0.753 and 0.775 Dice scores.

%
%\subsection{Performance on Large  Slice Reduction}
%

%\textbf{Results on large spacing displacement.}
\textbf{On large slice space.}  We further evaluate the robustness of Dual-PRNet to large slice space. Experiments were conducted on LPBA40, by reducing the slices of the moving volumes from 160$\times$192$\times$160 to 160$\times$24$\times$160. 
 Specifically, we preform the slice reduction on moving volumes by simply removing the slices to 160$\times$24$\times$160, and then interpolate the reduced volumes to the original size (160$\times$192$\times$160) with a spline interpolation (order=1). Then we perform our methods on the reduced-interpolated volumes which have the same size of the original moving volumes. By this way, we can verify different levels of slice reductions between the moving volume and the fixed volume, while keeping the fixed volumes unchanged.
During testing, the estimated final deformation field is applied to the labels of the moving volume using zero-order interpolation.
With a large reduction of slices from 192 to 24, our Dual-PRNet can still obtain a high average Dice score of 0.711, which even outperforms 0.683 of VoxelMorph~\citep{balakrishnan2018, balakrishnan2019} using the original non-reduced volumes. This demonstrates the strong robustness of our model against the large spacing displacements.

%\subsection{Performance on Detailed Structure}
\subsection{On Detailed Structure}
%\textbf{On detailed structure.} 
Compared with LPBA40 dataset, the Mindboggle101 is annotated with the cortical structure, which
contains more complicated anatomical structure of the brain, and often requires more accurate local detailed information to identify subtle difference.
%contains more fine detailed structure of the brain. 
Fig. \ref{fig:Mindboggle101} demonstrates the registration results on a number of MRI examples from the Mindboggle101.
As can be found, VoxelMorph does not provide accurate results on the detailed brain structure.
%while our methods have more favorable performance on these cases.
%
In addition, as demonstrated in ablation studies, our sequential pyramid registration with either PR modules or  PR$^{++}$  modules has larger improvements on the Mindboggle101 than that of the LPBA40. 
%We observed that images from the Mindboggle101  often have the more complicated anatomical structure of the brain, which often require more accurate local detailed information to identify subtle difference. 
Our sequential pyramid registration performs coarse-to-fine refinements of the deformation fields via sequential warping, which naturally aggregate more detailed information from multi-layer feature pyramids.  
%Furthermore, the local correlation features computed by our PR$^{++}$ modules enable Dual-PRNet$^{++}$ to better model the detailed structural information.
%
%With dual-stream convolutional features and pyramid registration fields, our Dual-PRNet obtains better results than VoxelMorph on the demonstrated cases. 
Furthermore, the enhanced Dual-PRNet$^{++}$ can achieve further higher performance by using the improved PR$^{++}$ modules, which are able to compute the local correlations explicitly and thus encodes more detailed structural information.
%First, the correlation layer compute the match cost between input feature maps, providing explicit correspondence for the deformation field estimation. Second, the feature fusion block allows the network to learn intrinsic correspondence on the feature space, resulting in stronger deep representation that are more discriminative.
The sequential warping allows the model to propagate the strong high-level context information gradually through the decoding layers, which enhances both the high-level semantic context and the local detailed structure.

We further perform more quantitative analysis. First, we can measure the fineness of the brain structures roughly by computing the ratio of labeled voxels to the total number of voxels in the volumes. The ratios of the labeled voxels on the Mindboggle101 and LPBA40 are 35.83\% and 60.02\% respectively, which demonstrate that the clinical regions defined in the Mindboggle101 are more fineness. Second, the difficulty of the brain structure on the two datasets can be demonstrated clearly by the performance of an Affine Registration, which has a 0.427 Avg Dice on the Mindboggle101  and a 0.669 Avg Dice on the LPBA40. Therefore,  brain structures presented in the Mindboggle101 are more challenging, and our Dual-PRNet++ achieved an over 75\% relative improvement, compared to 19\% relative improvement obtained on the LPBA40.

\subsection{On Cross-dataset Learning}
%\textbf{Cross-dataset evaluation.} 
We further evaluate the generalization capability of the proposed Dual-PRNet and Dual-PRNet$^{++}$ by conducting external cross-dataset validation. For example, we train on the LPBA40 and test on the Mindboggle101, and vice versa. Dual-PRNet$^{++}$ obtains an average Dice score of 0.788 on the LPBA40, and 0.739 on the Mindboggle101, which are slightly lower than the original performance: 0.798 and 0.748, respectively.  Similarly, the cross-data performance of Dual-PRNet are 0.747  and 0.581 on the two databases, compared to the original 0.778 and 0.631 respectively. Therefore, the cross-data performance of both Dual-PRNet and Dual-PRNet$^{++}$ are compared favorably against the original results of VoxelMorph (with 0.683 and 0.511 respectively), demonstrating the improved generalization ability of the proposed methods over different datasets.
%Visualization results on the cross-dataset are compared in SM.

\subsection{Limitations}

%\textbf{Limitations.} 
Dual-PRNet has its limitation by preforming sequential warping on deformation fields, which might result in an accumulation of interpolation artifacts.
Thus it yields the highest folding fraction. However, by integrating our PR$^{++}$ module, the folding fraction of deformation field on Dual-PRNet$^{++}$ can be reduced considerably, reaching a  higher performance compared favorably against VoxelMorph. 
However, we note that the current implementation of Dual-PRNet and Dual-PRNet++ do not have an explicit mechanism to regulate the amount of regularization as it is the case for FAIM \citep{kuang2019} or LapIRN \citep{mok2020large}. Nonetheless, inspired by these works, we note that adding a regularization term based on the determinant of the Jacobian matrix of the deformation is straightforward in our methods, and worth investigating in line with regularizing the solution where trading accuracy does not affect the downstream clinical task.
Besides, we further preformed our Dual-PRNet++ using additional supervision of segmentation masks, which can improve the performance from 0.748 to 0.752 on the Mindboggle101. However, this improvement is relatively limited when compared to that of VoxelMorph. \citep{balakrishnan2019}

Besides, in this work, we only performed our methods on 3D brain image (MRI) registration and segmentation. We expect that they can be further applied for or extended to more general 3D medical images, for registration or segmentation, but more experiments and evaluations are required, which can be kept as our future work.

\section{Conclusion}
We have presented our Dual-Stream Pyramid Registration Network (Dual-PRNet), with its extension, Dual-PRNet$^{++}$, for unsupervised 3D medical image registration. Our Dual-PRNet has two-stream architecture by design, which allows it to compute two convolutional  feature  pyramids  separately  from  two  input  volumes. Then sequential pyramid registration with a set of PR modules is proposed to estimate a sequence of registration fields, which can refine the learned pyramid features gradually in a coarse-to-fine manner via sequential warping. The PR module is further enhanced by computing local correlation features with further enhancement by residual convolutions, resulting in an enhanced Dual-PRNet$^{++}$. The proposed methods can be integrated into a 3D segmentation framework for joint registration and segmentation, where we demonstrate that it can greatly facilitate the segmentation task by accurately warping the voxel-level labels. Extensive experiments were conducted on  LPBA40 and Mindboggle101 databases, where the proposed Dual-PRNet$^{++}$ can outperform the stage-of-the-art methods  considerably  on unsupervised brain MRI registration.

%Based on the original two-stream 3D encoder-decoder and pyramid registration module, we integrated a correlation layer and feature fusion block into pyramid registration module to enforce the network to provide more corresponding information for deformation field. With these technical improvements, our method achieved impressive performance for brain MRI registration, and significantly outperformed recent approaches. Additionally, we employ our registration network as a data augmentor for segmentation network by providing large amounts of warped data. With the assistance of Dual-PRNet$^{++}$, the segmentation network improve the performance significantly with limited annotation, which extend the application of our work.

\section*{Acknowledgments}
We would like to thank Zhenlin Xu for sharing the implementation details of joint 3D segmentation framework (DeepAtlas), with image post-processing details on Mindboggle101 dataset.
% Acknowledgements Head style for the Acknowledgments heading.

%%Harvard
\bibliographystyle{model2-names}\biboptions{authoryear}
\bibliography{refs}

%\section*{Supplementary Material}

%Supplementary material that may be helpful in the review process should be prepared and provided as a separate electronic file. That file can then be transformed into PDF format and submitted along with the manuscript and graphic files to the appropriate editorial office.

\end{document}